\documentclass[table]{article}

\usepackage{graphicx}

\usepackage[round]{natbib}  
\usepackage{fullpage}
\usepackage{authblk}

\usepackage[utf8]{inputenc} 
\usepackage[T1]{fontenc}    
\usepackage{hyperref}       
\usepackage{url}            
\usepackage{booktabs}       
\usepackage{amsfonts}       
\usepackage{nicefrac}       
\usepackage{microtype}      

\usepackage{amsmath}
\usepackage{amsthm}
\usepackage{algorithm}
\usepackage{algorithmic}
\usepackage{xspace}
\usepackage[dvipsnames]{xcolor}
\usepackage{graphicx}
\usepackage{caption}
\usepackage{subcaption}
\usepackage{multirow}
\usepackage{tablefootnote}
\usepackage[usestackEOL]{stackengine}
\usepackage{colortbl} 
\usepackage{hhline}

\usepackage{enumitem}
\usepackage{Definitions}

\newcommand{\comment}[1]{}

\def\E{\mathbb{E}}
\def\R{\mathbb{R}}
\def\qpi{Q^{\pi}}

\def\mdp{\mathcal{M}}
\def\init{\mu_0}

\def\visitpi{d^\pi}

\def\visitrb{d^\Dset}

\DeclareMathOperator{\avgstep}{\rho}

\def\bellman{\mathcal{B}}

\def\Sset{S}
\def\Aset{A}

\def\Dset{\mathcal{D}}

\def\defeq{:=}
\renewcommand{\widehat}{\hat}

\def\rew{R}
\def\bellman{\gamma\cdot \mathcal{P}^{\pi}}

\def\bellmant{\gamma\cdot \mathcal{P}^{\pi}_{*}}

\def\bellmannog{\mathcal{P}^{\pi}}
\def\bellmantnog{\mathcal{P}^{\pi}_{*}}

\def\qvar{Q}
\def\dvar{d}

\def\initsamp{\substack{a_0\sim\pi(s_0) \\ s_0\sim\init}}

\def\ie{\emph{i.e.}\xspace}
\def\eg{\emph{e.g.}\xspace}

\def\piset{\{\pi_i\}_{i=1}^N}

\def\rank{\mathcal{O}}
\def\rloss{\mathcal{S}}

\renewcommand{\cite}{\citep}

\hypersetup{
colorlinks=true,
}

\renewcommand{\le}{\leqslant}

\renewcommand{\ge}{\geqslant}

\newcommand{\estname}{{BayesDICE}\xspace}
\newcommand{\Perm}{\mathrm{Perm}}
\newcommand{\zdist}[1]{Z(\cdot|\pi_{#1})}
\newcommand{\zmean}[1]{\overline{\rho}_{#1}}

\title{Offline Policy Selection under Uncertainty}

\author{
  $^*$Mengjiao Yang$^1$, \thanks{indicates equal contribution. Email: \texttt{\{sherryy, bodai, ofirnachum\}@google.com}.}
  Bo Dai$^1$, $^*$Ofir Nachum$^1$\\\vspace{-2mm}
  George Tucker$^1$, Dale Schuurmans$^{1,2}$\\ \vspace{3mm}
  $^1$Google Research, Brain Team \quad $^2$University of Alberta
}

\begin{document}

\date{}
\maketitle

\begin{abstract}
The presence of uncertainty in policy evaluation  significantly complicates the process of policy ranking and selection in real-world settings.
We formally consider \emph{offline policy selection} as learning preferences over a set of policy prospects given a fixed experience dataset. While one can select or rank policies based on point estimates of their policy values or high-confidence intervals, access to the full distribution over one's belief of the policy value enables more flexible selection algorithms under a wider range of downstream evaluation metrics. We propose \emph{\estname} for estimating this belief distribution in terms of posteriors of distribution correction ratios derived from stochastic constraints (as opposed to explicit likelihood, which is not available). 
Empirically, \estname is highly competitive to existing state-of-the-art approaches in confidence interval estimation.
More importantly, we show how the belief distribution estimated by BayesDICE may be used to rank policies with respect to any arbitrary downstream policy selection metric, and we empirically demonstrate that this selection procedure significantly outperforms existing approaches, such as ranking policies according to mean or high-confidence lower bound value estimates\footnote{Code available at~{\href{https://github.com/google-research/dice_rl}{https://github.com/google-research/dice\_rl}}.}.
\end{abstract}

\section{Introduction}

\emph{Off-policy evaluation} (OPE)~\citep{Precup00ET} in the context of reinforcement learning (RL) is often motivated as a way to mitigate risk in practical applications where deploying a policy might incur significant cost or safety concerns~\citep{thomas2015higha}. 
Indeed, by providing methods to estimate the value of a \emph{target policy} solely from a static \emph{offline} dataset of logged experience in the environment, OPE can help practitioners determine whether a target policy is or is not safe and worthwhile to deploy. 
Still, in many practical applications the ability to accurately estimate the online value of a specific policy is less of a concern than the ability to select or rank a set of policies (one of which may be the currently deployed policy).
This problem, related to but subtly different from OPE, is~\emph{offline policy selection}~\citep{doroudi2017importance,paine2020hyperparameter,kuzborskij2020confident}, and it often arises in practice.
For example, in recommendation systems, a practitioner may have a large number of policies trained offline using various hyperparameters, while cost and safety constraints only allow a few of those policies to be deployed as live experiments. Which policies should be chosen to form the small subset that will be evaluated online? 

This and similar questions are closely related to OPE, and indeed, the original motivations for OPE were arguably with offline policy selection in mind~\citep{Precup00ET,jiang2017theory}, the idea being that one can use estimates of the value of a set of policies to rank and then select from this set.
Accordingly, there is a rich literature of approaches for computing point estimates of the value of the policy~\citep{dudik2011doubly,JMLR:v14:bottou13a,jiang2015doubly,thomas2016data,NacChoDaiLi19,zhang2020gendice,uehara2020minimax,kallus2020double,yang2020offpolicy}. 
Because the offline dataset is finite and collected under a logging policy that may be different from the target policy, prior OPE methods also estimate high-confidence lower and upper bounds on a target policy's value~\citep{thomas2015higha,kuzborskij2020confident,JMLR:v14:bottou13a,hanna2016bootstrapping,feng2020accountable,dai2020coindice,kostrikov2020statistical}. 
These existing approaches may be readily applied to our recommendation systems example, by using either mean or lower-confidence bound estimates on each candidate policy to rank the set and picking the top few to deploy online.

However, this na\"{i}ve approach ignores crucial differences between the problem setting of OPE and the downstream evaluation criteria a practitioner prioritizes. 
For example, when choosing a few policies out of a large number of available policies, a recommendation systems practitioner may have a number of objectives in mind: The practitioner may strive to ensure that the policy with the overall highest groundtruth value is within the small subset of selected policies (akin to top-$k$ precision).
Or, in scenarios where the practitioner is sensitive to large differences in achieved value, a more relevant downstream metric may be the difference between the largest groundtruth value within the $k$ selected policies compared to the groundtruth of the best possible policy overall (akin to top-$k$ regret).
With these or other potential offline policy selection metrics, it is far from obvious that ranking according to OPE estimates is ideal~\citep{doroudi2017importance}. 

The diversity of potential downstream metrics in offline policy selection presents a challenge to any algorithm that yields a point estimate for each policy. Any one approach to computing point estimates will necessarily be sub-optimal for some policy selection criteria.
To circumvent this challenge, we propose to compute a \emph{belief distribution} over groundtruth values for each policy.
Specifically, with the posteriors for the distribution over value for each policy calculated, one can use a straightforward procedure that takes estimation uncertainty into account to rank the policy candidates according to arbitrarily complicated downstream metrics. 
While this belief distribution approach to offline policy selection is attractive, it also presents its own challenge: how should one estimate a distribution over a policy's value in the pure offline setting?

In this work, we propose \emph{Bayesian Distribution Correction Estimation (BayesDICE)} for off-policy estimation of a belief distribution over a policy's value. BayesDICE works by estimating
posteriors over correction ratios for each state-action pair
(correcting for the distribution shift between the off-policy data and the target policy's on-policy distribution). A belief distribution of the policy's value may then be estimated by averaging these correction distributions over the offline dataset, weighted by rewards. In this way, BayesDICE builds on top of the state-of-the-art DICE point estimators~\citep{NacChoDaiLi19,zhang2020gendice,yang2020offpolicy}, while uniquely leveraging posterior regularization to satisfy chance constraints in a Markov decision process (MDP). As a preliminary experiment, we show that BayesDICE is highly competitive to existing frequentist approaches when applied to confidence interval estimation. More importantly, we demonstrate BayesDICE's application in offline policy selection under different utility measures on a variety of discrete and continuous RL tasks. Among other findings, our policy selection experiments suggest that, while the conventional wisdom focuses on using lower bound estimates to select policies (due to safety concerns)~\citep{kuzborskij2020confident}, policy ranking based on the lower bound estimates does not always lead to lower (top-$k$) regret. Furthermore, when other metrics of policy selection are considered, such as top-$k$ precision, being able to sample from the posterior enables significantly better policy selection
than only having access to the mean or confidence bounds of the estimated policy values.

\section{Preliminaries}
We consider an infinite-horizon Markov decision process
(MDP)~\citep{puterman1994markov} denoted as $\mdp = \langle\Sset, \Aset, \rew, T, \init, \gamma\rangle$, which consists of a state space, an action space, a deterministic reward function,\footnote{For simplicity, we restrict our analysis to deterministic rewards, and extending our methods to stochastic reward scenarios is straightforward.} a transition probability function, an initial state distribution, and a discount factor $\gamma\in (0, 1]$. In this setting, a policy $\pi(a_t|s_t)$ interacts with the environment
starting at $s_0 \sim \init$ and receives a scalar reward $r_t=\rew(s_t, a_t)$ as the environment transitions into a new state $s_{t+1} \sim T(s_t, a_t)$ at each timestep $t$. The value of a policy is defined as 
\begin{equation}\label{eq:policy_value_def}
\textstyle
\rho\rbr{\pi} \defeq \rbr{1-\gamma}\EE_{s_0, a_t, s_t}\sbr{\sum_{t=0}^\infty \gamma^t r_t}.
\end{equation}

\subsection{Offline policy selection}
We formalize the \emph{offline policy selection} problem as providing a ranking $\rank\in\Perm([1,N])$ over a set of \emph{candidate} policies $\piset$ given only a
\emph{fixed} dataset $\Dset=\{x^{(j)}\defeq(s_0^{(j)}, s^{(j)},a^{(j)},r^{(j)},s^{\prime(j)})\}_{j=1}^n$
where $s_0^{(j)}\sim\init$, $(s^{(j)},a^{(j)})\sim\visitrb$
are samples of an unknown distribution $\visitrb$,
$r^{(j)}=\rew(s^{(j)},a^{(j)})$,
and $s^{\prime(j)}\sim T(s^{(j)},a^{(j)})$.\footnote{This tuple-based representation of the dataset is for notational and theoretical convenience, following~\citet{dai2020coindice,kostrikov2020statistical}, among others. In practice, the dataset is usually presented as finite-length trajectories $\{(s_0^{(j)},a_0^{(j)},r_0^{(j)},s_1^{(j)},\dots)\}_{j=1}^m$, and this can be processed into a dataset of finite samples from $\init$ and from $\visitrb\times \rew\times T$. For mathematical simplicity, we assume that the dataset is sampled i.i.d. This is a common assumption in the OPE literature~\citep{uehara2020minimax} and may be relaxed in some cases by assuming a fast mixing time~\citep{NacChoDaiLi19}.}
One approach to the offline policy selection problem is to first characterize the \emph{value} of each policy (Eq.~\ref{eq:policy_value_def}, also known as the normalized
per-step reward) via OPE under some \emph{utility} function $u(\pi)$ that leverages a point estimate (or lower bound) of the policy value; i.e., 
\begin{equation*}
    \rank \leftarrow \mathrm{ArgSortDescending}(\{u(\pi_i)\}_{i=1}^N).
\end{equation*}

\subsection{Selection evaluation}\label{sec:ops}
A proposed ranking $\rank$ will eventually be evaluated according to how well its policy ordering aligns with the policies' groundtruth values. In this section, we elaborate on potential forms of this evaluation score.

To this end, let us denote the groundtruth distribution of returns of policy $\pi_i$ by $\zdist{i}$. In other words, $\zdist{i}$ is a distribution over $\R$ such that
\begin{equation}
    z\sim \zdist{i} \equiv \left [ z:=(1-\gamma)\sum_{t=0}^\infty \gamma^t\cdot\rew(s_t,a_t)~;~ s_0\sim \init, a_t\sim\pi_i(s_t), s_{t+1}\sim T(s_t,a_t) \right].
\end{equation}
Note that $\EE_{\zdist{i}}\left[z\right] = \rho(\pi_i)$.

As part of the offline policy selection problem, we are given a \emph{ranking score} $\rloss$ that is a function of a proposed ranking $\rank$ and groundtruth policy statistics $\{\zdist{i}\}_{i=1}^N$.
The ranking score $\rloss$ can take on many forms and is application specific; \eg,
\begin{itemize}
	\item{\textbf{top-$k$ precision}: This is an \emph{ordinal} ranking score. The ranking score considers the top $k$ policies in terms of groundtruth means $\rho(\pi_i)$ and returns the proportion of these which appear in the top $k$ spots of $\rank$. 
	}
	\item{\textbf{top-$k$ accuracy}: Another ordinal ranking score, this score considers the top-$k$ policies in sorted order in terms of groundtruth means $\rho(\pi_i)$ and returns the proportion of these which appear in the same ordinal location in $\rank$. 
	}
	\item{\textbf{top-$k$ correlation}: Another ordinal ranking score, this represents the Pearson correlation coefficient between the ranking of top-$k$ policies in sorted order in terms of groundtruth means $\rho(\pi_i)$ and the truly best top-$k$ policies.
	}	
	\item{\textbf{top-$k$ regret}: This is a \emph{cardinal} ranking score. This score respresents the difference in groundtruth means $\rho(\pi_i)$ between the overall best policy -- \ie, $\max_i \rho(\pi_i)$ -- and the best policy among the top-$k$ ranked policies -- \ie, $\max_{i\in[1,k]}\rho(\pi_{\rank[k]})$.
	}
	\item{\textbf{Beyond expected return}: One may define the above ranking scores in terms of statistics of $\zdist{i}$ other than the groundtruth means $\rho(\pi_i)$. For example, in safety-critical applications, one may be concerned with the variance of the policy return. Accordingly, one may define CVaR analogues to top-$k$ precision and regret.
	}
\end{itemize}
For simplicity, we will restrict our attention to ranking scores which only depend on the average return of $\pi_i$. To this end, we will use $\zmean{i}$ as shorthand for $\rho(\pi_i)$ and assume that the ranking score $\rloss$ is a function of $\rank$ and $\{\zmean{i}\}_{i=1}^N$.

\subsection{Ranking score simulation from the posterior}
It is not clear whether ranking according to vanilla OPE (either mean or confidence based) is ideal for any of the ranking scores above, including, for example,
top-$1$ regret in the presence of uncertainty.
However, if one has access to an approximate belief distribution over the policy's values, there is a simple sampling-based approach that can be used to find a near-optimal ranking (optimality depending on how accurate the belief distribution is) with respect to an arbitrary specified downstream ranking score, and we elaborate on this procedure here.

First, note that if we have access to the true groundtruth policy values $\{\zmean{i} \}_{i=1}^N$, and the ranking score function $\rloss$, we can calculate the score value of \emph{any} ranking $\rank$ and find the ranking $\rank^*$ that optimizes this score. However, we are limited to a finite offline dataset and the full return distributions are unknown. In this offline setting, we propose to instead compute a belief distribution $q(\{\zmean{i} \}_{i=1}^N)$, and then we can optimize over the expected ranking score $\mathbb{E}_{q}\left[\rloss(\rank, \{\zmean{i} \}_{i=1}^N)\right]$ as shown in Algorithm~\ref{algo:offline-select}. 
This algorithm simulates realizations of the groundtruth values $\{\zmean{i}\}_{i=1}^N$ by sampling from the belief distribution $q(\{\zmean{i} \}_{i=1}^N)$, and in this way estimates the expected realized ranking score $\rloss$ over all possible rankings $\rank$.
As we will show empirically, matching the selection process (the $\rloss$ used in Algorithm~\ref{algo:offline-select}) to the downstream ranking score naturally leads to improved performance. The question now becomes how to effectively learn a belief distribution over $\{\zmean{i} \}_{i=1}^N$.

\begin{minipage}{0.5\textwidth}
\centering
\includegraphics[width=1.\textwidth]{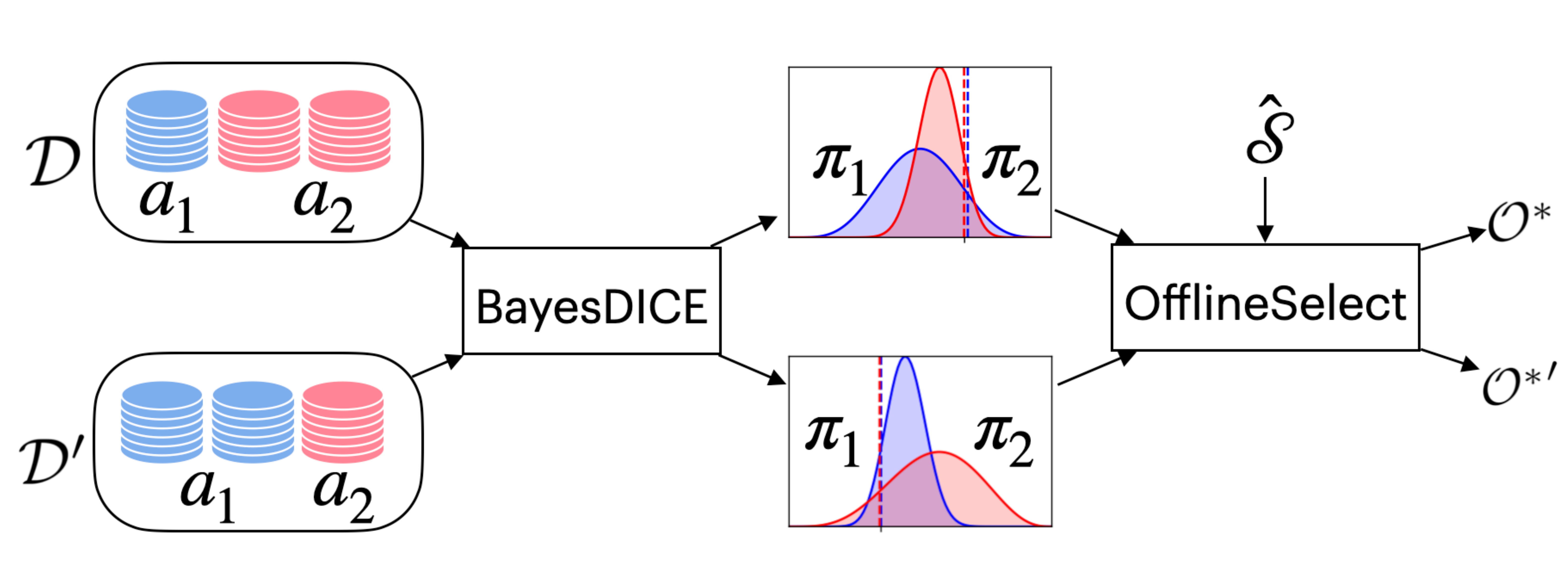}
\captionof{figure}{The belief distributions of $\zmean{1}$ and $\zmean{2}$ depend on the uncertainty induced from the finite offline data ($\Dcal$ and $\Dcal'$). 
A user might prefer $\pi_2$ only if $p(\zmean{2} < \zmean{1}) < \xi$ (a choice of $\mathcal{S}$). 
OPE based on mean point estimates would select $\pi_2$ in either case as $\zmean{2}$ has the greater mean.
Sampling from the posterior belief in \texttt{OfflineSelect} allows simulation of any ranking score under $\rloss$, aligning policy selection with the user's choice of $\rloss$.}
\label{fig:data-to-posterior}
\end{minipage}
\hfill\begin{minipage}{0.45\textwidth}
\begin{algorithm}[H]
\caption{OfflineSelect}
\begin{algorithmic}
\STATE \textbf{Inputs}~Posteriors $q(\{\zmean{i}\}_{i=1}^N)$, ranking score $\hat{\rloss}$
\STATE Initialize $\rank^*; L^*$ \hfill\texttt{\footnotesize $\rhd$ Track best score}
\FOR {$\rank$ in \texttt{Perm}($[1,...,N]$)}
	\STATE $L = 0$
	\FOR {$j = 1$ to $n$}
		\STATE sample $\{\hat{\rho}_i^{(j)}\}_{i=1}^N \sim q(\{\zmean{i}\}_{i=1}^N)$
		\STATE \texttt{\footnotesize $\rhd$ Sum up sample scores}
		\STATE $L = L + \hat{\rloss}(\{\hat{\rho}_i^{(j)}\}_{i=1}^N, \rank)$
	\ENDFOR
	\IF {$L < L^*$}
	\STATE \texttt{\footnotesize $\rhd$ Update best ranking/score}
		\STATE $L^* = L$; $\rank^* = \rank$
	\ENDIF
\ENDFOR
\RETURN $\rank^*, L^*$
\end{algorithmic}
\label{algo:offline-select}
\end{algorithm}
\end{minipage}

\section{BayesDICE}
To learn a belief distribution over $\{\zmean{i} \}_{i=1}^N$,
we pursue a Bayesian approach to
infer an approximate posterior distribution given prior beliefs.
While model-based Bayesian approaches exist (\eg,~\citep{deisenroth2011pilco} and variants~\citep{parmas2018pipps}), they typically suffer from compounding error, so a model-free approach is preferable.
However, Bayesian inference is challenging in this model-free scenario because the likelihood function is not easy to compute, as it is defined over infinite horizon returns.

Therefore, we first investigate several approaches to representing policy value, before identifying a novel posterior estimator that is computationally attractive and can support
a broad range
of ranking scores for downstream tasks. 

\subsection{Policy Ranking Score Representation}
In practice, the downstream task of ranking or selecting policy candidates might require more than the value expectation, but also other properties of the policy value distribution. To ensure that the necessary distribution properties are computable, we 
first consider the class of ranking scores we would like to support:
\begin{itemize}
	\item {\bf Offline}: Since we focus on ranking policies given only \emph{offline} data, the ranking score should not depend on on-policy samples.  
	\item {\bf Flexible}: Since the downstream task may utilize different ranking scores,
	the representation of the policy value should be sufficient to support their efficient computation.
\end{itemize}

With these considerations in mind, we review ways to represent
 the value of a policy $\pi$.
Define
$Q^\pi\rbr{s, a} = \textstyle\EE\sbr{\sum_{t=0}^\infty \gamma^t \rew(s_t,a_t) | s_0=s, a_0=a}$
and
\begin{align*}
d^\pi\rbr{s, a} &= \textstyle\rbr{1 - \gamma}\sum_{t=0}^\infty \gamma^t d_t^\pi\rbr{s, a},
\quad\mbox{with} 
\\
d_t^\pi\rbr{s, a} &= \Pb\rbr{s_t = s, a_t = a|s_0\sim\mu_0, \forall i<t, a_i\sim \pi\rbr{\cdot|s_i}, s_{i+1}\sim T\rbr{\cdot|s_i,a_i}}
,
\end{align*}
which are the \emph{state-action} values and \emph{stationary visitations} of $\pi$.
These satisfy the recursions
{\small
\begin{eqnarray}
\label{eq:bellman-q}
\textstyle\qpi(s,a) 
\!\!&\!\!=\!\!&\!\! 
\rew(s,a) + \bellman \qpi(s,a),
    \text{ where }  \bellmannog \qvar(s,a) \defeq \E_{s'\sim T(s,a),a'\sim\pi(s')}[\qvar(s',a')];
\\
\label{eq:bellman-d}
\textstyle\visitpi(s,a) 
\!\!&\!\!=\!\!&\!\! 
(1-\gamma)\init(s)\pi(a|s) + \bellmant\visitpi(s,a),
    \text{ where }  \bellmantnog \dvar(s,a) \defeq \pi(a|s) \textstyle\sum_{\tilde{s},\tilde{a}} T(s|\tilde{s},\tilde{a})\dvar(\tilde{s},\tilde{a}).
    \qquad
\end{eqnarray}
}
From these identities,
the policy value can be expressed in two equivalent ways:
\begin{align}\label{eq:policy_value_primal_dual}
  \avgstep(\pi) &= (1-\gamma) \cdot \E_{\initsamp}[\qpi(s_0,a_0)] \\
  &= \E_{(s,a)\sim\visitpi}[r(s,a)]. 
  \label{eq:policy_value_dual}
\end{align}
Current OPE methods are generally based on one of the representations \eqref{eq:policy_value_def}, \eqref{eq:policy_value_primal_dual} or
\eqref{eq:policy_value_dual}. For example,
importance sampling (IS) estimators
\citep{Precup00ET,Murphy01MM,dudik2011doubly} are  based on~\eqref{eq:policy_value_def}; LSTDQ~\citep{lagoudakis2003least} is a representative algorithm for fitting $Q^\pi$ and thus based on~\eqref{eq:policy_value_primal_dual}; the recent DICE algorithms~\citep{nachum2020reinforcement,yang2020offpolicy} estimate the stationary density ratio $\zeta\rbr{s, a}\defeq \frac{d^\pi\rbr{s, a}}{\visitrb}$ so that $\rho\rbr{\pi} = \EE_{\visitrb}\sbr{\zeta\cdot r}$, and are thus based on~\eqref{eq:policy_value_dual}. 

Among the three strategies, the third is the most promising in our scenario.
First, IS suffers from an exponential growth in variance~\citep{liu2018breaking} and further requires knowledge of the behavior policy. 
In contrast, the functions $Q^\pi$ and $d^\pi$ are duals \citep{nachum2020reinforcement,yang2020offpolicy},
and share common \emph{behavior-agnostic} and minimax properties~\citep{uehara2020minimax}, However, estimation of $Q^\pi$ assumes a ranking score with a linear dependence on $R\rbr{s, a}$, and therefore, even if we estimate $Q^\pi$  accurately, it is still impossible to evaluate ranking scores that involve $\rbr{1-\gamma}\EE\sbr{\sum_{t=0}^\infty \gamma^t \sigma(r_t)}$ such that $\sigma(\cdot):\RR\rightarrow \RR$ is a nonlinear function (unless one learns a different $Q$ function for each possible ranking score, which may be computationally expensive).  By contrast, ranking scores with such nonlinear components can be easily computed from the stationary density ratio as $\EE_{\visitrb}\sbr{\zeta\cdot\sigma\rbr{r}}$.
Given these considerations, the estimator via stationary density ratio satisfies both requirements: it enjoys statistical advantages in the offline setting and is flexible for downstream ranking score calculation. 
Therefore, we focus on a Bayesian estimator for $\zeta^\pi$ next. 

\subsection{Stationary Ratio Posterior Estimation}
Recall that
to apply a simple Bayesian approach to infer the posterior of $\zeta^\pi$, one requires a $\log$-likelihood function, but such a quantity is not readily calculable in our scenario from the given data.
Therefore, we develop an alternative, computationally tractable approach
by considering an optimization view of Bayesian inference under a chance constraint, 
which allows us to derive the posterior over a set of stochastic equations. 

Let $f\rbr{\cdot}$ denote a non-negative convex function with $f(0)$ achieving the minimum $0$, \eg, $f(x) = x^\top x$.
Also let $\Delta_d\rbr{s, a} \defeq  (1-\gamma)\init(s)\pi(a|s) + \bellmant d(s,a) - d\rbr{s, a} $.
Starting with~\eqref{eq:policy_value_primal_dual} 
we reduce the $\abr{S}\abr{A}$ many constraints for the stationary distribution of $\pi$ to a single feature-vector-based constraint for $\zeta$: 
\begin{eqnarray}\label{eq:stationary_metric}
&&\Delta_d\rbr{s, a} = 0, \quad \forall (s, a)\in S\times A\Rightarrow \inner{\phi}{\Delta_d} = 0\\
&\Rightarrow& f\rbr{\inner{\phi}{\Delta_d}} = 0 \Rightarrow \max_{\beta\in \Hcal_\phi} \beta^\top\inner{\phi}{\Delta_d} - f^*\rbr{\beta} = 0\\
&\Rightarrow&{\max_{\beta\in \Hcal_\phi} \EE_{d^\Dcal}\sbr{ \zeta\rbr{s, a}\cdot \beta^\top\rbr{\gamma \phi(s', a')- \phi\rbr{s, a}}}} + \rbr{1 - \gamma}\EE_{\mu_0\pi}\sbr{\beta^\top\phi} - f^*\rbr{\beta} = 0,
\end{eqnarray}

where $\Hcal_\phi$ denotes the bounded Hilbert space with the feature mappings $\phi$, $\visitrb$ denotes the distribution generating the empirical experience,
and we have used Fenchel duality in the middle step. 
The function $\phi\rbr{\cdot, \cdot}: S\times A\rightarrow \RR^m$ is a feature mapping,
with $m$ possibly infinite. 
Then the condition $\inner{\phi}{\Delta_d} =0$ can be understood as matching the two distributions $(1-\gamma)\init(s)\pi(a|s) + \bellmant d(s,a)$ and $d\rbr{s, a}$ in terms of their embeddings~\citep{smola2007hilbert}, which is a generalization of the approximation methods in~\citep{de2003linear,Lakshminarayanan17}. 
In particular, when $\abr{S}\abr{A}$ is finite and we set $\phi(s, a) = {\delta_{s, a}}$,
where $\delta_{s, a}\in\cbr{0, 1}^{\abr{S}\abr{A}}$ is an indicator vector with a single $1$  at position $(s, a)$ and $0$ otherwise, we are matching the distributions pointwise. The feature map $\phi\rbr{s, a}$ can also be set to general reproducing kernel $k\rbr{(s, a), \cdot}\in \RR^\infty$. As long as the kernel $k\rbr{\cdot, \cdot}$ is characteristic, the embeddings will match if and only if the distributions are identical almost surely~\citep{sriperumbudur2011universality}.

Given that the experience was collected by some other means, \ie, $\Dcal\sim\visitrb$, the constraint for $\zeta$ in~\eqref{eq:stationary_metric} might not hold exactly. Therefore, we consider a feasible set $\zeta\in \cbr{\zeta: {\ell\rbr{\zeta,\Dcal}} \le \epsilon}$ where
\begin{equation}
\textstyle
\ell\rbr{\zeta,\Dcal}\defeq{\max_{\beta\in \Hcal_\phi} \widehat\EE_{\Dcal}\sbr{ \zeta\rbr{s, a}\cdot \beta^\top\rbr{\gamma \phi(s', a')- \phi\rbr{s, a}} - f^*\rbr{\beta}}}+ \rbr{1 - \gamma}\EE_{\mu_0\pi}\sbr{\beta^\top\phi}
.
\end{equation}
Note that $\ell\rbr{\zeta}\ge 0$ since $\Hcal_\phi$ is symmetric. We expect the posterior of $\zeta$, $q\rbr{\zeta}$, to concentrate most of its mass on this set and balance the prior. Formally, this means 
\begin{equation}
\min_{q}\,\, KL\rbr{q||p} - \lambda \xi,\quad \st\,\,\PP_q\rbr{\ell\rbr{\zeta}\le \epsilon}\ge \xi,
\end{equation}
where the chance constraint considers the probability of the feasibility of $\zeta$ under the posterior. This formulation can be equivalently rewritten as 
\begin{eqnarray}\label{eq:bayes_prob}
\min_{q}\,\,&&KL\rbr{q||p} - \lambda \PP_q\rbr{\ell\rbr{\zeta}\le \epsilon}
\end{eqnarray}
Then, by applying Markov's inequality, \ie, 
$\PP_q\rbr{\ell\rbr{\zeta}\le \epsilon} = 1 - \PP_q\rbr{\ell\rbr{\zeta}\ge \epsilon} \ge 1 - \frac{\EE_q\sbr{\ell\rbr{\zeta}}}{\epsilon}$,
we can obtain an upper bound on~\eqref{eq:bayes_prob} as
\begin{eqnarray}
&&
\min_{q}\,\,KL\rbr{q||p} + \frac{\lambda}{\epsilon} \EE_q\sbr{\ell(\zeta,\Dcal)} \label{eq:bayesdice_I}\\
&=&
\min_{q(\zeta)}\max_{q(\beta|\zeta)}KL\rbr{q||p} + \frac{\lambda}{\epsilon} \EE_{q\rbr{\zeta}q\rbr{\beta|\zeta}}\Big[\widehat\EE_{\Dcal}\sbr{ \zeta\rbr{s, a}\cdot \beta^\top\rbr{\gamma \phi(s', a')- \phi\rbr{s, a}} - f^*\rbr{\beta}} 
\nonumber\\
&&
+ \rbr{1 - \gamma}\EE_{\mu_0\pi}\sbr{\beta^\top\phi}\Big] 
\label{eq:bayesdice_II},
\end{eqnarray}
where the equality follows by interchangeability~\citep{shapiro2014lectures,dai2017learning}. 
We amortize the optimization for $\beta$ w.r.t. each $\zeta$  to a distribution $q\rbr{\beta|\zeta}$ to reduce the computational effort. 

Due to the space limitation, we postpone the discussion about the important properties of \estname, including the parametrization of the posteriors, the variants of \estname for undiscounted MDP and alternatives of the $\log$-likelihoods, and the connections to the vanilla Bayesian stochastic processes, to~\appref{app:more_bayesdice}. Please refer the details there. 

Finally, note that with the posterior approximation for $\zeta_i$, denoting the estimate for candidate policy $i$, we can draw posterior samples of $\bar{\rho}_i$ by drawing a sample $\zeta_i \sim q(\zeta_i)$ and computing $\hat{\rho}_i = \frac{1}{n}\sum_{(s, a, r) \in \Dset}\zeta_i(s, a)r$. This defines a posterior distribution over $\bar{\rho}_i$ and we further assume that the distributions are independent for each policy, so $q(\{\bar{\rho}_i\}_{i=1}^N) = \prod_i q(\bar{\rho}_i)$. This defines the necessary inputs for \texttt{OfflineSelect} to determine a ranking of the candidate policies.

\section{Related work}\label{sec:related_work}
We categorize the relevant related work into three categories: offline policy selection, off-policy evaluation, and Bayesian inference for policy evaluation.

\paragraph{Offline policy selection}
The decision making problem we formalize as offline policy selection is a member of a set of problems in RL referred to as~\emph{model selection}.
Previously, this term has been used to refer to state abstraction selection~\citep{jiang2017theory,jiang2015abstraction} as well as learning algorithm and feature selection~\citep{foster2019model,pacchiano2020model}.
More relevant to our proposed notion of policy selection are a number of previous works which use model selection to refer to the problem of choosing a near-optimal $Q$-function from a set of candidate approximation functions~\citep{fard2010pac,farahmand2011model,irpan2019off,xie2020batch}. In this case, the evaluation metric is typically defined as the $L_\infty$ norm of difference of $Q$ versus the state-action value function of the optimal policy $Q^*$.
While one can relate this evaluation metric to the sub-optimality (i.e., regret) of the policy induced by the $Q$-function, we argue that our proposed policy selection problem is both more general -- since we allow for the use of policy evaluation metrics other than sub-optimality -- and more practically relevant -- since in many practical applications, the policy may not be expressible as the argmax of a $Q$-function. 
Lastly, the offline policy selection problem we describe is arguably a formalization of the problem approached in~\citet{paine2020hyperparameter} and referred to as~\emph{hyperparameter selection}. In contrast to this previous work, we not only formalize the decision problem, but also propose a method to directly optimize the policy selection evaluation metric.
Offline policy selection has also been studied by~\citet{doroudi2017importance}, which considers what properties a point estimator should have in order for it to yield good rankings in terms of a notion of ranking score referred to as \emph{fairness}.

\paragraph{Off-policy evaluation}
Off-policy evaluation (OPE) is a highly active area of research. While the original motivation for OPE was in the pursuit of policy selection~\citep{Precup00ET,jiang2017theory}, the field has historically almost exclusively focused on the related but distinct problem of estimating the online value (accumulated rewards) of a single target policy. In addition to a plethora of techniques for providing point estimates of this groundtruth value~\citep{dudik2011doubly,JMLR:v14:bottou13a,jiang2015doubly,thomas2016data,kallus2020double,NacChoDaiLi19,zhang2020gendice,yang2020offpolicy}, there is also a growing body of literature that uses frequentist principles to derive high-confidence lower bounds for the value of a policy~\citep{JMLR:v14:bottou13a,thomas2015highb,hanna2016bootstrapping,kuzborskij2020confident,feng2020accountable,dai2020coindice,kostrikov2020statistical}.
As our results demonstrate, ranking or selecting policies based on either their estimated mean or lower confidence bounds can at times be sub-optimal, depending on the evaluation criteria. 

\paragraph{Bayesian inference for policy evaluation}
Our proposed method for policy selection relies on Bayesian principles to estimate a posterior distribution over the groundtruth policy value. While many Bayesian-inspired methods have been proposed for policy optimization~\citep{deisenroth2011pilco, parmas2018pipps}, especially in the context of exploration~\citep{houthooft2016vime,dearden2013model,kolter2009near}, relatively few have been proposed for policy evaluation. In one instance,~\citet{fard2010pac} derive PAC-Bayesian bounds on estimates of the Bellman error of a candidate $Q$-value function. In contrast to this work, we use our BayesDICE algorithm to estimate a distribution over target policy value, and this distribution allows us to directly optimize arbitrary downstream policy selection metrics.

\section{Experiments}
We empirically evaluate the performance of \estname on confidence interval estimation (which can be used for policy selection) and offline policy selection under linear and neural network posterior parametrizations on tabular -- Bandit, Taxi~\citep{dietterich1998maxq}, FrozenLake~\citep{brockman2016openai} -- and continuous-control -- Reacher~\citep{brockman2016openai} -- tasks. As we show below, \estname outperforms existing methods for confidence interval estimation, producing accurate coverage while maintaining tight interval width, suggesting that BayesDICE achieves accurate posterior estimation, being robust to approximation errors and potentially misaligned Bayesian priors in practice.
Moreover, in offline policy selection settings, matching the selection algorithm (Algorithm~\ref{algo:offline-select}) to the ranking score (enabled by the estimating the posterior) shows clear advantages over ranking based on point estimates or confidence intervals on a variety of ranking scores. See~\appref{app:exp} for additional results and implementation details.

\subsection{Confidence interval estimation}\label{exp:ci}
Before applying BayesDICE to policy selection, we evaluate the BayesDICE approximate posterior by computing the accuracy of the confidence intervals it produces. We compare \estname against a known set of confidence interval estimators based on concentration inequalities. To compute these baselines, we first use weighted (\ie, self-normalized) per-step importance sampling~\citep{thomas2016data} to compute a policy value estimate for each logged trajectory. These trajectories provide a finite sample of value estimates. We use self-normalized importance sampling in MDP environments since it has been found to yield better empirical results despite being biased~\citep{liu2018breaking,NacChoDaiLi19}. We then use empirical \emph{Bernstein's} inequality~\citep{thomas2015highb}, bias-corrected \emph{bootstrap}~\citep{thomas2015higha}, and \emph{Student's t-test} to derive lower and upper high-confidence bounds on these estimates. We further consider Bayesian Deep Q-Networks (BDQN)~\cite{azizzadenesheli2018efficient} with an average empirical reward prior in the function approximation setting, which applies Bayesian linear regression to the last layer of a deep Q-network to learn a distribution of Q-values. Both \estname and BDQN output a distribution of parameters, from which we conduct Monte Carlo sampling and use the resulting samples to compute a confidence interval at a given confidence level.

\begin{figure}[h]
\centering
  \begin{subfigure}{1.\columnwidth}
    \includegraphics[width=1.\linewidth]{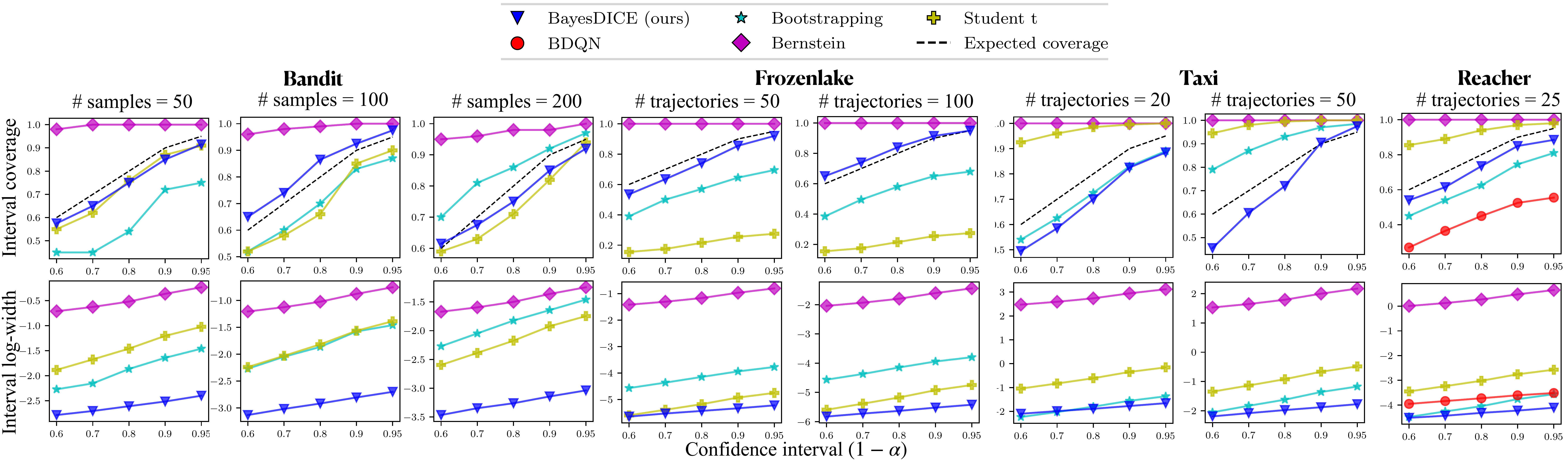}
  \end{subfigure}
  \caption{Confidence interval estimation on Bandit, FrozenLake, Taxi, and Reacher. The $y$-axis shows the empirical coverage and median log-interval width across $200$ trials. BayesDICE exhibits near true coverage while maintaining narrow interval width, suggesting an accurate posterior approximation.}
\label{fig:ci}  
\end{figure}

We plot the empirical coverage and interval width at different confidence levels in~\figref{fig:ci}. To compute the empirical \emph{interval coverage}, we conduct $200$ trials with randomly sampled datasets. The interval coverage is the proportion of the $200$ intervals that contains the true value of the target policy. The \emph{interval log-width} is the median of the log width of the $200$ intervals. As shown in \figref{fig:ci}, \estname's coverage closely follows the intended coverage (black dotted line), while maintaining narrow interval width across all tasks considered.
This suggests that \estname's posterior estimation is highly accurate, being robust to approximation errors and potentially misaligned Bayesian priors in practice. 

\subsection{Policy selection}
\begin{figure}[h]
\centering
  \begin{subfigure}{1.\columnwidth}
  \centering
    \includegraphics[width=0.75\linewidth]{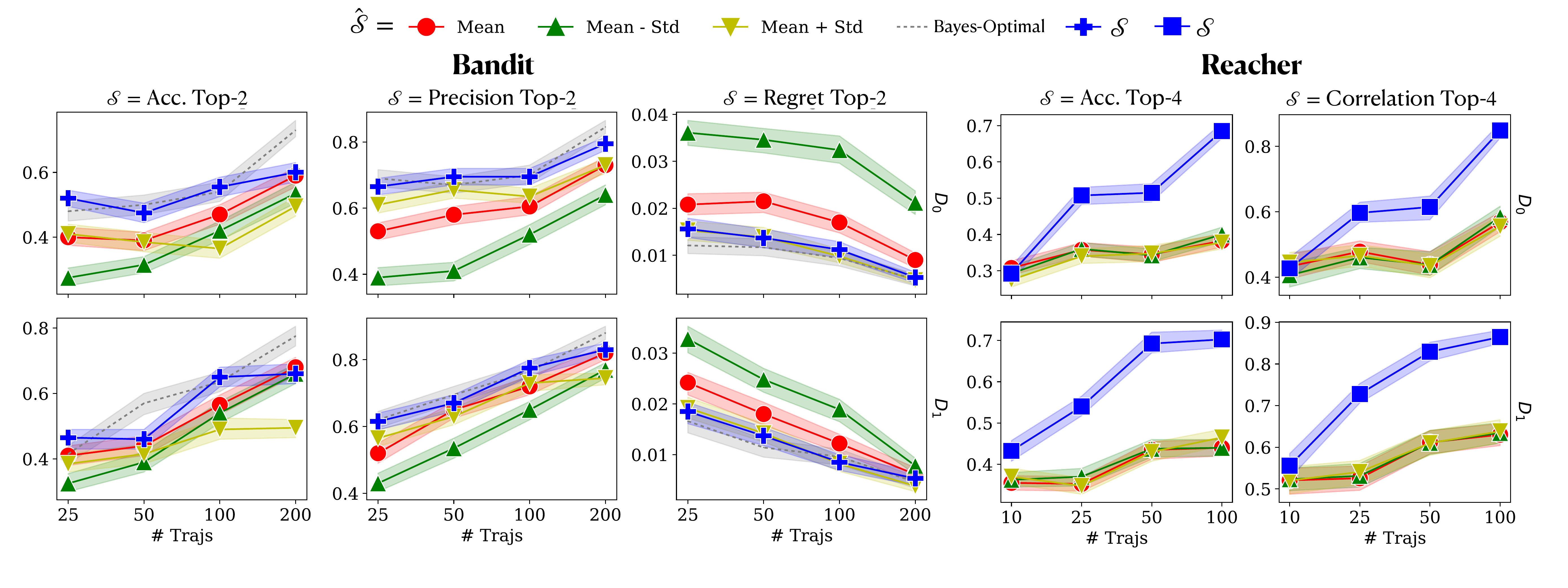}    
  \end{subfigure} 
  \caption{Policy selection using top-$k$ ranking scores compared to mean/confidence ranking approaches on two-armed Bandit and Reacher. 
  In these experiments, we fix the posterior to the one approximated by BayesDICE and evaluate different $\hat{\rloss}$ used in Algorithm~\ref{algo:offline-select} to compute a policy ranking.
  We find that using $\hat{\rloss}=\rloss$ (\ie, aligning the ranking score in posterior simulation with the groundtruth evaluation) results in better performance than simple point estimates.
  Interestingly, the lower-bound point estimate almost always performs worse than the mean or the upper bound.
  }
\label{fig:ablation}  
\end{figure}

Next, we demonstrate the benefit of matching the policy selection criteria to the ranking score in offline policy selection. Our evaluation is based on a variety of cardinal and ordinal ranking scores defined in~\secref{sec:ops}. 
We begin by considering the use of Algorithm~\ref{algo:offline-select} with BayesDICE-approximated posteriors. By keeping the BayesDICE posterior fixed, we focus our evaluation on the performance of Algorithm~\ref{algo:offline-select}.
We plot the groundtruth performance of this procedure applied to Bandit and Reacher in~\figref{fig:ablation}. These figures compare using different $\hat{\rloss}$ to rank the policies according to Algorithm~\ref{algo:offline-select} across different downstream ranking scores $\rloss$.
We find that aligning the criteria $\hat{\rloss}$ used in Algorithm~\ref{algo:offline-select} with the downstream ranking score $\rloss$ is empirically the best approach ($\hat{\rloss} = \rloss$). In contrast, using point estimates such as $\texttt{Mean}$ or $\texttt{Mean} \pm \texttt{Std}$ can yield much worse downstream performance. 
We also see that in the Bandit setting, where we can analytically compute the Bayes-optimal ranking, using aligned ranking scores in conjunction with BayesDICE-approximated posteriors achieves near-optimal performance.

\begin{figure}[h]
\centering 
   \includegraphics[width=0.8\linewidth]{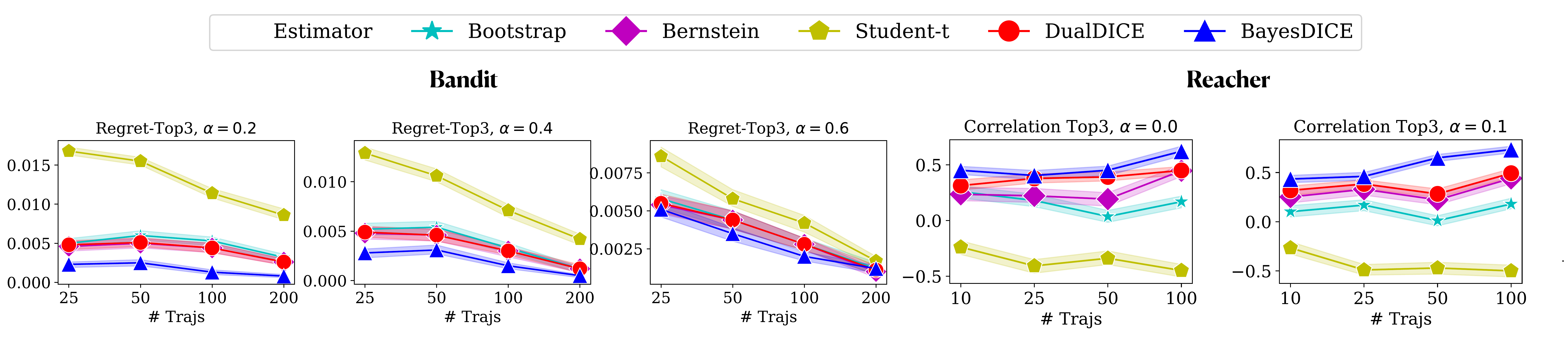}
  \caption{Policy selection evaluation under correlation and regret at top-$k$ in two-armed Bandit (left) and  Reacher (right) compared to other methods using point estimate (DualDICE) or high-confidence lower bounds.
  Please see \appref{app:exp} for more results with respect to other downstream metrics.
  }
\label{fig:selection_bandit}  
\end{figure}

Having established BayesDICE's ability to compute accurate posterior distributions as well as the benefit of appropriately aligning the ranking score used in Algorithm~\ref{algo:offline-select}, we compare BayesDICE to state-of-the-art OPE methods in policy selection.
In these experiments, we use Algorithm~\ref{algo:offline-select} with posteriors approximated by BayesDICE and $\hat{\rloss} = \rloss$.
We compare the use of BayesDICE in this way to ranking via point estimates of DualDICE~\citep{NacChoDaiLi19} and other confidence-interval estimation methods introduced in~\secref{exp:ci}. 
We present results in Figure~\ref{fig:selection_bandit}, in terms of top-$k$ regret and correlation on bandit and reacher across different sample sizes and behavior data. BayesDICE outperforms other methods on both tasks. See additional ranking results in~\appref{app:exp}.

\section{Conclusion}
In this paper, we formally defined the offline policy selection problem, and proposed~\estname to first estimate posterior distributions of policy values before using a simulation-based procedure to compute an optimal policy ranking. Empirically, \estname not only provides accurate belief distribution estimation, but also shows excellent performance in policy selection tasks.

\subsubsection*{Acknowledgments}
We thank members of the Google Brain team for helpful discussions.

\bibliography{main}
\bibliographystyle{plainnat}

\clearpage
\newpage
\clearpage
\newpage

\appendix
\onecolumn

\begin{appendix}

\thispagestyle{plain}
\begin{center}
{\huge Appendix}
\end{center}

\section{More Discussions on BayesDICE}\label{app:more_bayesdice}

In this section, we provide more details about \estname. 

\paragraph{Remark (parametrization of $q\rbr{\zeta}$ and $q\rbr{\beta|\zeta}$):} We parametrize both $q\rbr{\zeta}$ (and the resulting $q\rbr{\beta|\zeta}$) as Gaussians with the mean and variance approximated by a multi-layer perceptron (MLP),~\ie: $\zeta = \text{MLP}_w(s, a) + \sigma_{w'}\xi, \,\,\xi\sim\Ncal(0, 1)$. $w$ and $w'$ denote the parameters of the MLP.

\paragraph{Remark (connection to Bayesian inference for stochastic processes):} Recall the posterior can be viewed as the solution to an optimization~\citep{Zellner88,ZhuCheXin14,DaiHeDaiSon16}, 
\[
q\rbr{\zeta|\Dcal} = \argmin_{q\in\Pcal}\,\, \inner{q\rbr{\zeta}}{\log p\rbr{\zeta, \Dcal}} + KL\rbr{q\rbr{\zeta}||p\rbr{\zeta}}
\]
The~\eqref{eq:bayesdice_I} is equivalent to define the $\log$-likelihood proportion to $\ell\rbr{\zeta, \Dcal}$, which is a stochastic process, including Gaussian process~($\Gcal\Pcal$) by setting $f^*\rbr{\beta} = \frac{1}{2}\beta^\top\beta$. Specifically, plug $f\rbr{\beta} = \frac{1}{2}\beta^\top \beta$ back into~\eqref{eq:bayesdice_I}, we have $\beta^* = \widehat\EE_{\Dcal}\sbr{\zeta\rbr{s, a}\cdot\rbr{\gamma \phi(s', a')- \phi\rbr{s, a}}} + \rbr{1 - \gamma}\EE_{\mu_0\pi}\sbr{\phi}$, resulting the optimization
\begin{equation}
\min_{q}KL\rbr{q||p} + \frac{\lambda}{\epsilon} \EE_q{\EE_{\mu_0\pi}\widehat\EE_{\Dcal}\sbr{ \zeta\rbr{s_1, a_1}^\top k\rbr{\rbr{s_1, a_1, s_1', a_1'}, \rbr{s_2, a_2, s_2', a_2'}}\zeta\rbr{s_2,a_2}  }},
\end{equation}
with the kernel $k\rbr{x_1, x_2} \defeq \rbr{\gamma \phi(s_1', a_1')- \phi\rbr{s_1, a_1}}^\top \rbr{\gamma \phi(s_2', a_2')- \phi\rbr{s_2, a_2}} + \rbr{1 - \gamma}^2{\phi\rbr{s_1^0, a_1^0}^\top \phi\rbr{s_1^0, a_1^0}} + 2\rbr{1 - \gamma} \phi\rbr{{s_1^0, a_1^0}}^\top\rbr{\gamma \phi(s_2', a_2')- \phi\rbr{s_2, a_2}}$, which is a $\Gcal\Pcal$. Obviously, with different choices of $f^*\rbr{\cdot}$, the~\estname framework is far beyond $\Gcal\Pcal$.

Although the $\Gcal\Pcal$ has been applied for RL~\citep{engel2003bayes,ghavamzadeh2016bayesian,azizzadenesheli2018efficient}, they all focus on prior on value function; while \estname considers general stochastic processes likelihood, including $\Gcal\Pcal$, for the stationary ratio modeling, which as we justified is more flexible for different selection criteria in downstream tasks. 

\paragraph{Remark (auxilary constraints and undiscounted MDP):} As~\citet{yang2020offpolicy} suggested, the non-negative and normalization constraints are important for optimization. We exploit positive neuron to ensure the non-negativity of the mean of the $q\rbr{\zeta}$. For the normalization, we consider the chance constraints $\PP\rbr{\rbr{\widehat\EE_\Dcal\rbr{\zeta}-1}^2\le \epsilon_1}\ge\xi_1$. By applying the same technique, it leads to extra term $\frac{\lambda_1}{\epsilon_1}\EE_{q}\sbr{\max_{\alpha\in\RR}{\alpha\cdot\widehat\EE_\Dcal\sbr{\zeta - 1}}}$ in~\eqref{eq:bayesdice_I}.

With the normalization condition introduced, the proposed~\estname is ready for undiscounted MDP by simply setting $\gamma = 1$ in~\eqref{eq:bayesdice_I} together with the above extra term for normalization. 

\paragraph{Remark (variants of $\log$-likelihood):} We apply the Markov's inequality to~\eqref{eq:bayes_prob} for the upper bound~\eqref{eq:bayesdice_I}. In fact, the optimization with chance constraint has rich literature~\citep{BenElgNem09}, where plenty of surrogates can be derived with different safe approximation. For example, if the $q$ is simple, one can directly calculate the CDF for the probability $\PP_q\rbr{\ell\rbr{\zeta}\le \epsilon}$; or one can also exploit different probability inequalities to derive other surrogates, \eg, condition value-at-risk, \ie, 
\begin{equation}
\min_{q}\,\,KL\rbr{q||p} + {\lambda}\inf_t\sbr{t + \frac{1}{\epsilon}\EE_q\sbr{\ell\rbr{\zeta} - t} }_+,
\end{equation}
and Bernstein approximation~\cite{NemSha07}. These surrogates lead to better approximation to the chance probability $\PP_q\rbr{\ell\rbr{\zeta}\le \epsilon}$ with the extra cost in optimization.

\section{BayesDICE for Exploration vs. Exploitation Tradeoff}\label{app:bayes_dice}

In main text, we mainly consider exploiting BayesDICE for estimating various ranking scores for both discounted MDP and undiscounted MDP. In fact, with the posterior of the stationary ratio computed, we can also apply it for better balance between exploration vs. exploitation for policy optimization.

Instead of selecting from a set of policy candidates, the policy optimization is considering all feasible policies and selecting optimistically. Specifically, the feasibility of the stationary state-action distribution can be characterized as
\begin{equation}
	\sum_a d\rbr{s, a} = \rbr{1 - \gamma}\mu_0 + \Pcal_* d\rbr{s}, \quad \forall s\in S,
\end{equation}
where $\Pcal_* d\rbr{s}\defeq \sum_{\sbar, \abar}T\rbr{s|\sbar, \abar}d\rbr{\sbar, \abar}$. Apply the feature mapping for distribution matching, we obtain the constraint for $\zeta\cdot \pi$ with $\zeta\rbr{s, a}\defeq \frac{d\rbr{s}}{d^\Dcal(s,a)}$ as
\begin{equation}
	\max_{\beta\in\Hcal_\phi}\,\,\beta^\top\EE_{d^\Dcal}\sbr{\sum_a\rbr{\zeta\rbr{s,a}\pi\rbr{a|s}}\phi\rbr{s} - \gamma \rbr{\zeta\rbr{s, a}\pi\rbr{a|s}}\phi\rbr{s'}} + \rbr{1 - \gamma}\EE_{\mu_0}\sbr{\beta^\top\phi} -f^*\rbr{\beta} = 0.
\end{equation}
Then, we have the posteriors for all valid policies should satisfies
\begin{equation}
\lambda \PP_q\rbr{\ell\rbr{\zeta\cdot\pi, \Dcal}\le \epsilon}\ge \xi,
\end{equation}
with $\ell\rbr{\zeta\cdot\pi, \Dcal}\defeq \max_{\beta\in\Hcal_\phi}\,\,\beta^\top\widehat\EE_{\Dcal}\sbr{\sum_a\rbr{\zeta\rbr{s,a}\pi\rbr{a|s}}\phi\rbr{s} - \gamma \rbr{\zeta\rbr{s, a}\pi\rbr{a|s}}\phi\rbr{s'}} + \rbr{1 - \gamma}\EE_{\mu_0}\sbr{\beta^\top\phi} -f^*\rbr{\beta}$.
Meanwhile, we will select one posterior from among these posteriors of all valid policies optimistically, \ie, 
\begin{eqnarray}\label{eq:exp_bayesdice}
\max_{q\rbr{\zeta}q\rbr{\pi}}&& \EE_q\sbr{U\rbr{\tau, r, \Dcal}} + \lambda_1 \xi - \lambda_2KL\rbr{q\rbr{\zeta}q\rbr{\pi}||p\rbr{\zeta, \pi}}\\
\st&& \PP_q\rbr{\ell\rbr{\zeta\cdot\pi, \Dcal}\le \epsilon}\ge \xi
\end{eqnarray}
where $\EE_q\sbr{U\rbr{\tau, r, \Dcal}}$ denotes the optimistic policy score to capture the upper bound of the policy value estimation. For example, the most widely used one is
$$
\EE_q\sbr{U\rbr{\tau, r, \Dcal}} = \EE_q\widehat\EE_\Dcal\sbr{\tau\cdot r} + \lambda_u \EE_q\sbr{\rbr{\widehat\EE_{\Dcal}\sbr{\tau\cdot r} - \EE_q\widehat\EE_{\Dcal}\sbr{\tau\cdot r}}^2 },
$$
where the second term is the empirical variance and usually known as one kind of ``exploration bonus''.

Then the whole algorithm is iterating between solving~\eqref{eq:exp_bayesdice} and use the obtain policy collecting data into $\Dcal$ in~\eqref{eq:exp_bayesdice}. 

This Exploration-\estname follows the same philosophy of~\citet{osband2019deep,o2018uncertainty} where the variance of posterior of the policy value is taken into account for exploration. However, there are several significant differences: {\bf i),} the first and most different is the modeling object,~\citet{osband2019deep,o2018uncertainty} is updating with $Q$-function, while we are handling the dual representation; {\bf ii),} \estname is compatible with arbitary nonlinear function approximator, while~\citet{osband2019deep,o2018uncertainty} considers tabular or linear functions; {\bf iii),} \estname is considering infinite-horizon MDP, while~\citet{osband2019deep,o2018uncertainty} considers fixed finite-horizon case.  Therefore, the exploration with~\estname pave the path for principle and practical exploration-vs-exploitation algorithm. The regret bound  is out of the scope of this paper, and we leave for future work. 

\section{Experiment details and additional results}\label{app:exp}
\subsection{Environments and policies.}
\paragraph{Bandit.} We create a Bernoulli two-armed bandit with binary rewards where $\alpha$ controls the proportion of optimal arm ($\alpha = 0$ and $\alpha = 1$ means never and always choosing the optimal arm respectively). Our selection experiments are based on $5$ target policies with $\alpha = [0.75, 0.8, 0.85, 0.9, 0.95]$.
\paragraph{Reacher.} We modify the Reacher task to be infinite horizon, and sample trajectories of length 100 in the behavior data. To obtain different behavior and target policies, We first train a deterministic policy from OpenAI Gym~\citep{brockman2016openai} until convergence, and define various policies by converting the optimal policy into a Gaussian policy with optimal mean with standard deviation $0.4 - 0.3\alpha$. Our selection experiments are based on $5$ target policies with $\alpha = [0.75, 0.8, 0.85, 0.9, 0.95]$.
\subsection{Details of neural network implementation}
We parametrize the distribution correction ratio as a Gaussian using a deep neural network for the continuous control task. Specifically, we use feed-forward networks with two hidden-layers of $64$ neurons each and ReLU as the activation function. The networks are trained using the Adam optimizer ($\beta_1=0.99$, $\beta_2=0.999$) with batch size $2048$.

\subsection{Additional experimental results}
\begin{figure}[h]
\centering 
   \includegraphics[width=0.9\linewidth]{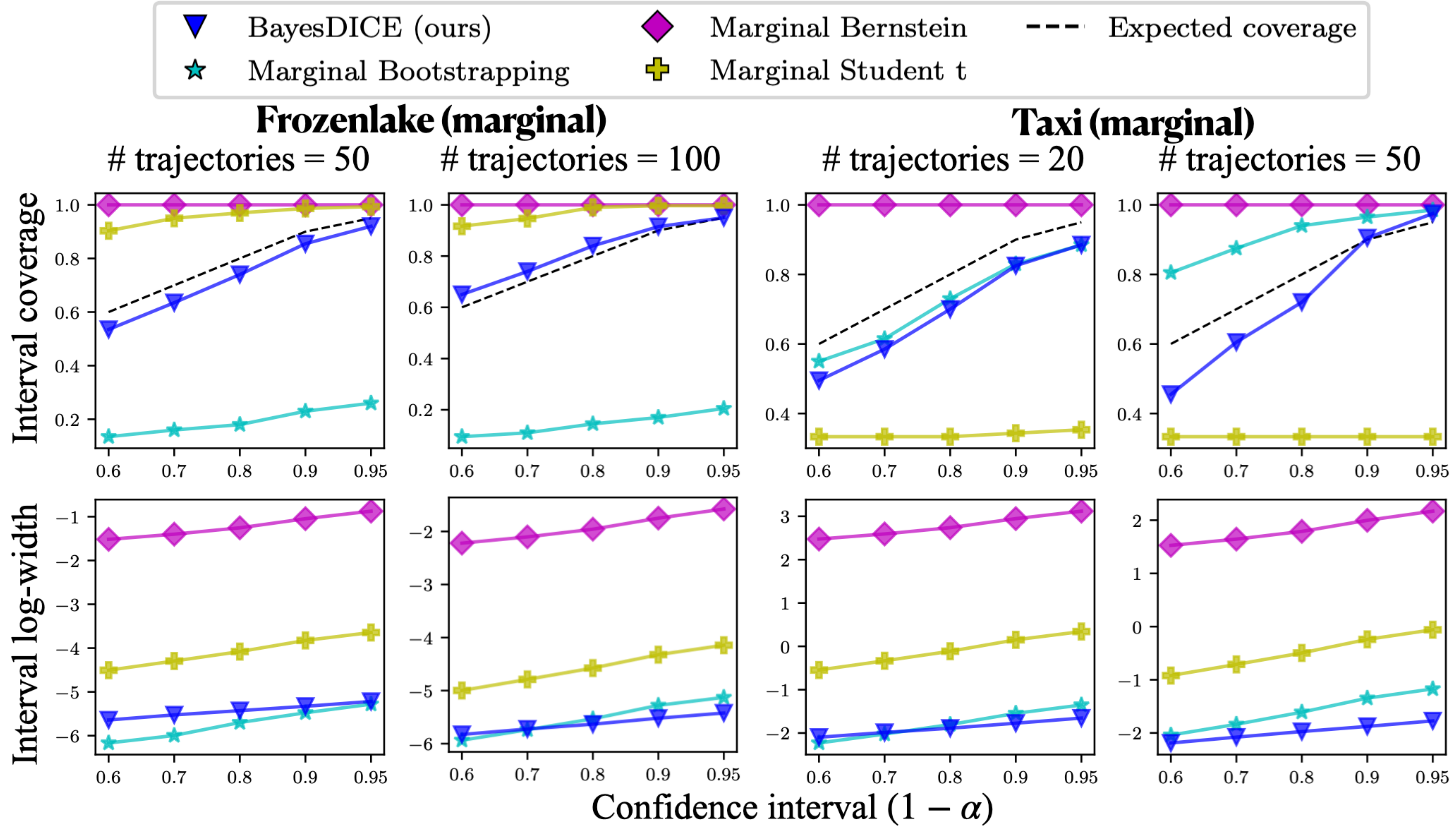}
  \caption{Confidence interval estimation with baselines computed from marginalized importance sampling.}
\label{fig:ci_marginal}  
\end{figure}

\begin{figure}[h]
\centering 
  \begin{subfigure}{0.8\columnwidth}
    \includegraphics[width=1.\linewidth]{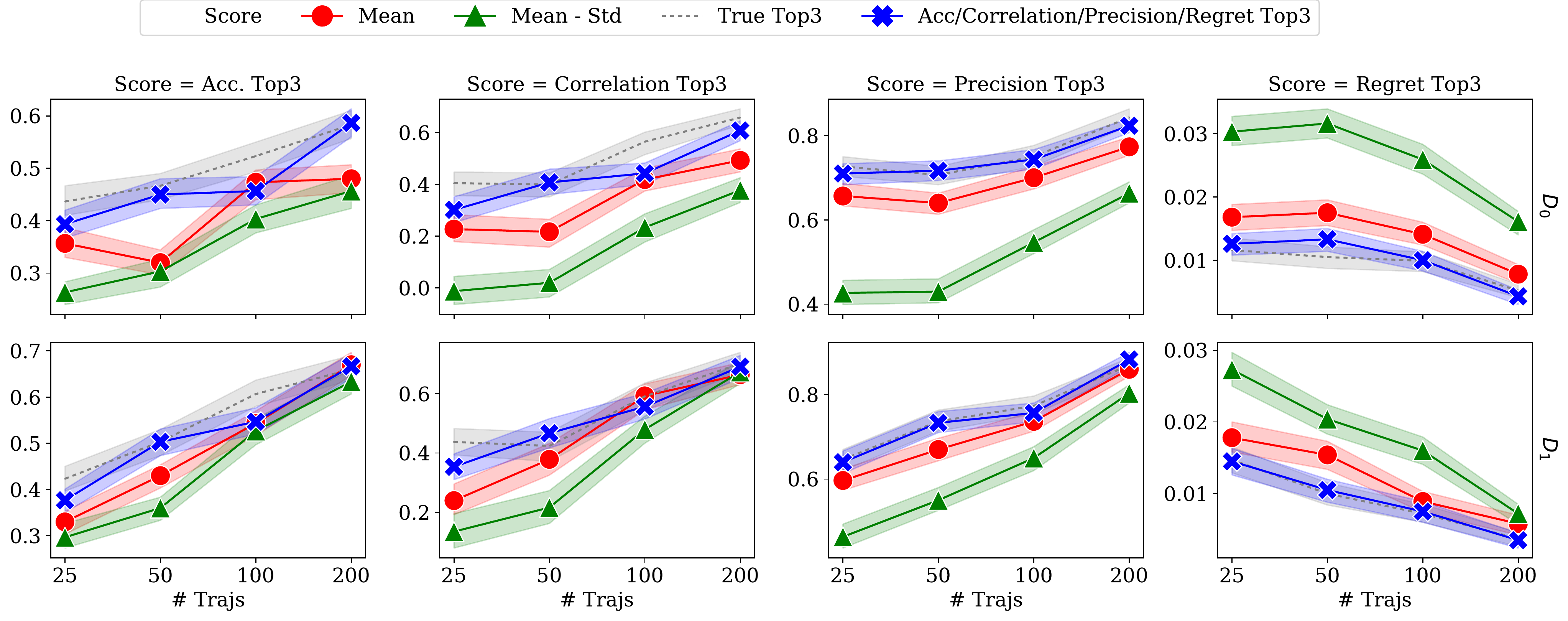}  
  \end{subfigure} 
  \begin{subfigure}{0.8\columnwidth}
    \includegraphics[width=1.\linewidth]{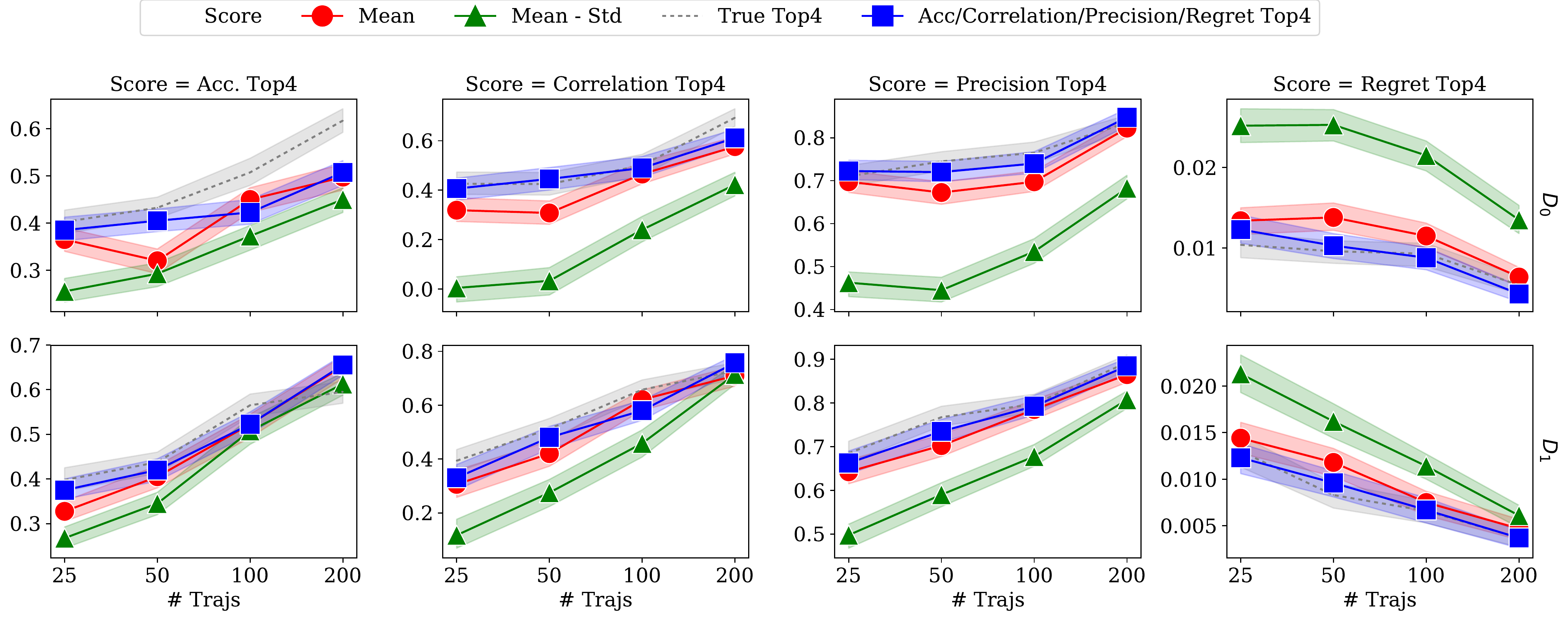}  
  \end{subfigure}   
  \caption{Additional $k$ values for top-$k$ ranking on bandit. Ranking results based on Algorithm~\ref{algo:offline-select} (blue lines) always perform better than using mean or high-confidence lower bound.}
\end{figure}
\begin{figure}[h]
\centering 
  \begin{subfigure}{0.8\columnwidth}
    \includegraphics[width=1.\linewidth]{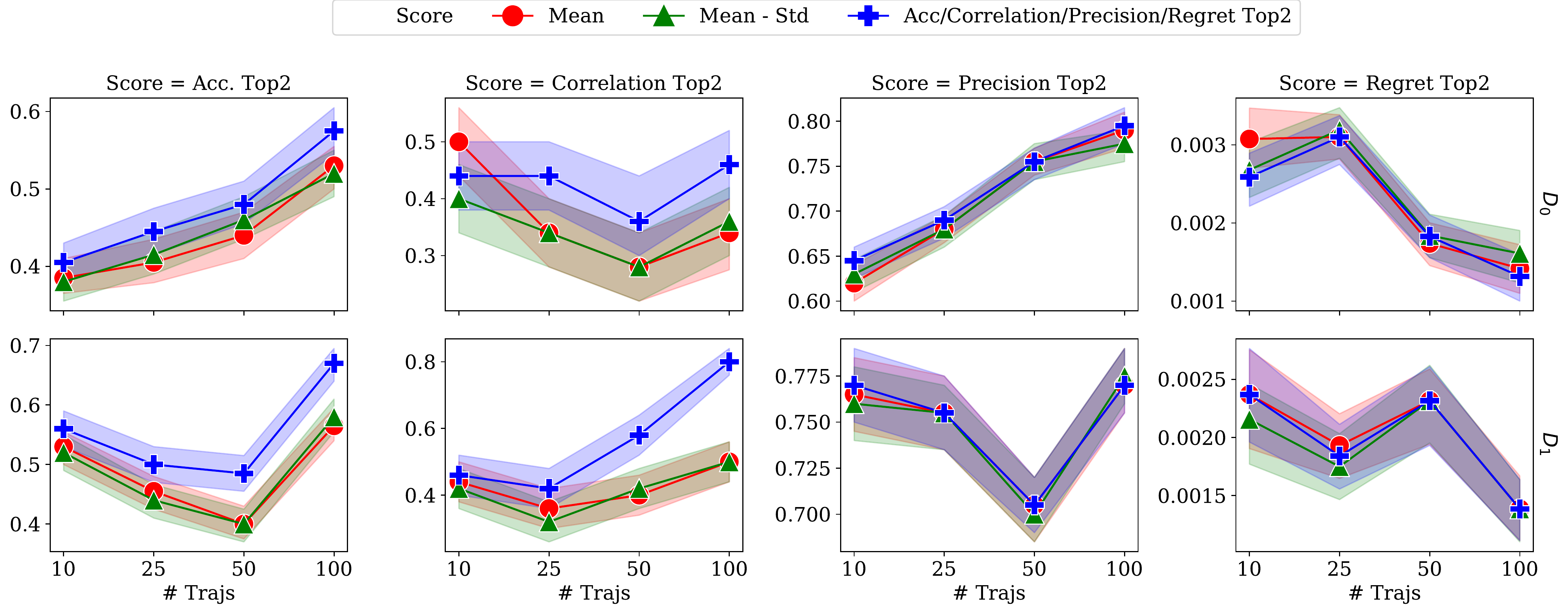}  
  \end{subfigure} 
  \begin{subfigure}{0.8\columnwidth}
    \includegraphics[width=1.\linewidth]{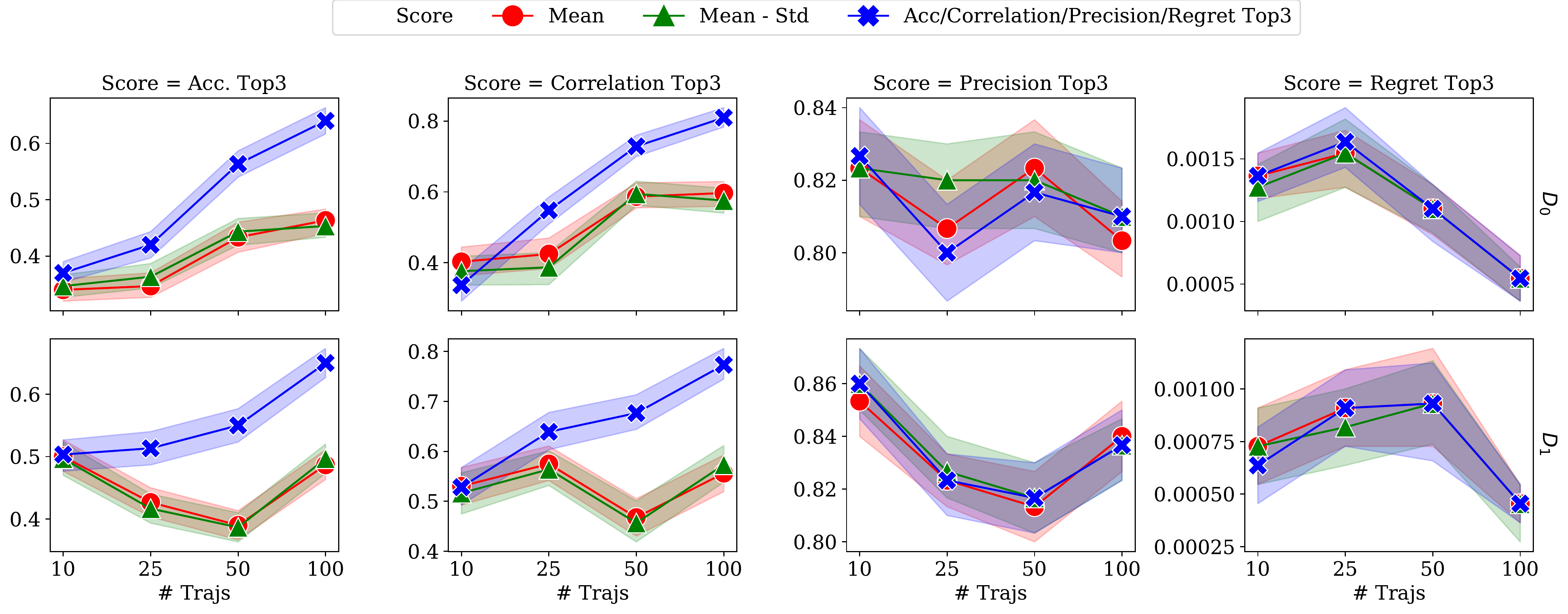}  
  \end{subfigure}   
  \caption{Additional $k$ values for top-$k$ ranking on reacher and additional scores (precision and regret). Ranking results based on Algorithm~\ref{algo:offline-select} (blue lines) generally perform much better than using mean or high-confidence lower bound for top-$k$ accuracy and correlation. Precision and regret are similar between posterior samples and the mean/confidence bound based ranking.}
\end{figure}

\begin{figure}[h]
\centering 
   \includegraphics[width=0.8\linewidth]{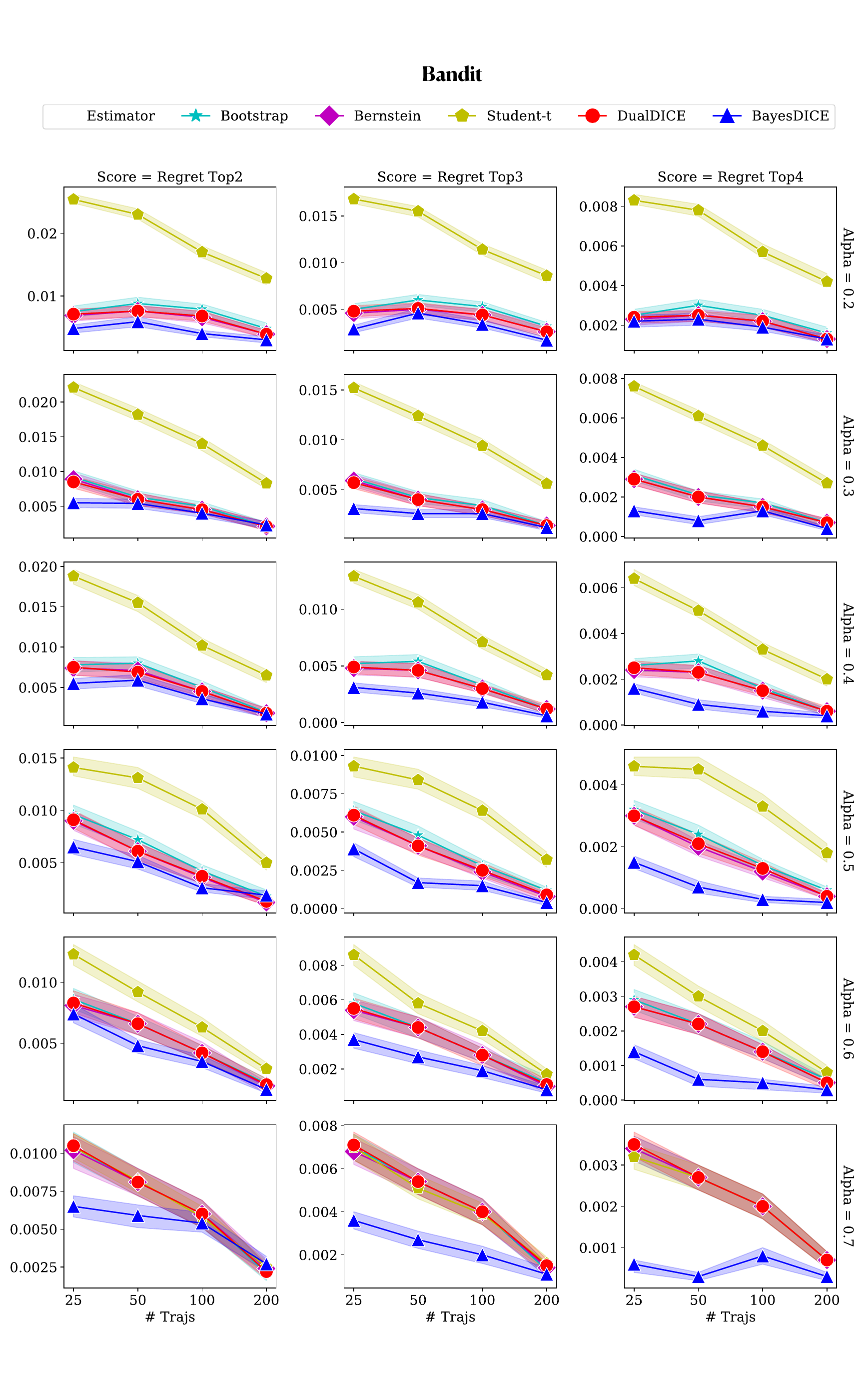}
  \caption{Improved regret using BayesDICE across all trajectory lengths, behavior data, and top-$k$ values considered for the bandit task.}
\label{fig:bandit_regret_all}  
\end{figure}
\begin{figure}[h]
\centering 
   \includegraphics[width=0.8\linewidth]{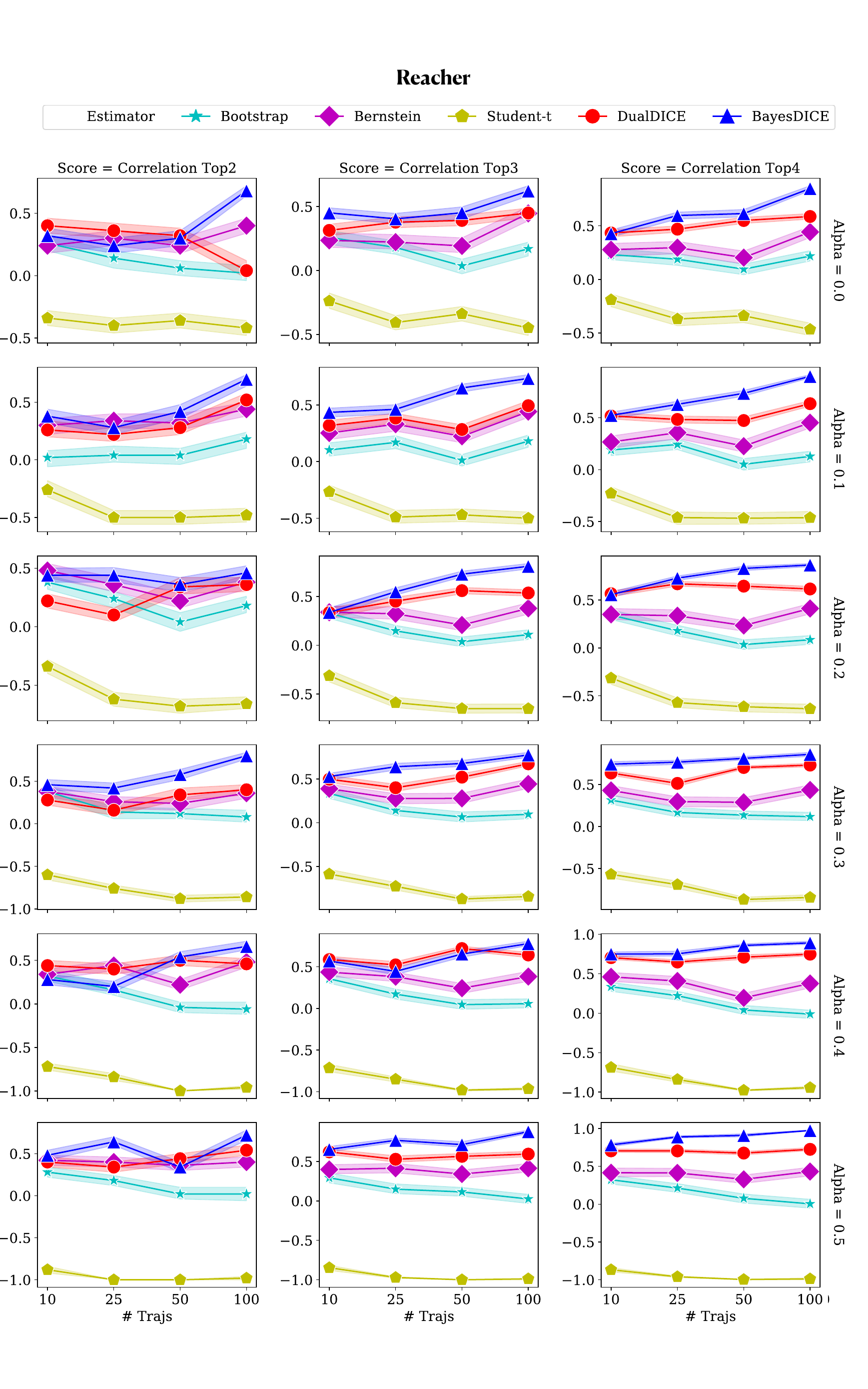}
  \caption{Improved correlation using BayesDICE across all trajectory lengths, behavior data, and top-$k$ values considered for the reacher task.}
\label{fig:reacher_corr_all}  
\end{figure}

\end{appendix}

\end{document}